\documentclass[10pt,twocolumn,letterpaper]{article}

\usepackage{3dv}
\usepackage{times}
\usepackage{epsfig}
\usepackage{graphicx}
\usepackage{amsmath}
\usepackage{amssymb}

\usepackage[breaklinks=true,colorlinks,bookmarks=false]{hyperref}

\usepackage{comment}
\usepackage{color}

\usepackage{epsfig}
\usepackage{graphicx}
\usepackage{amsmath}
\usepackage{amssymb}
\usepackage{booktabs}
\usepackage{pifont}

\usepackage{multirow}
\usepackage[percent]{overpic}
\usepackage{subcaption}

\definecolor{gold}{rgb}{1.0, 0.874, 0}
\definecolor{silver}{rgb}{0.77,0.77,0.77}
\definecolor{brown}{rgb}{0.95, 0.678, 0.4}

\newcommand{\gold}[1]{\colorbox{gold}{\textbf{#1}}}
\newcommand{\silver}[1]{\colorbox{silver}{\textbf{#1}}}
\newcommand{\bronze}[1]{\colorbox{brown}{\textbf{#1}}}

\newcommand\blfootnote[1]{%
  \begingroup
  \renewcommand\thefootnote{}\footnote{#1}%
  \addtocounter{footnote}{-1}%
  \endgroup
}

\usepackage{array}
\usepackage{color, colortbl}
\usepackage{pifont}

\usepackage[accsupp]{axessibility}  % Improves PDF readability for those with disabilities.

\definecolor{bad}{rgb}{1,0.7,0.7}

\newcolumntype{;}{!{\vrule width 2pt}}

\usepackage[capitalize]{cleveref}
\crefname{section}{Sec.}{Secs.}
\Crefname{section}{Section}{Sections}
\Crefname{table}{Table}{Tables}
\crefname{table}{Tab.}{Tabs.}

\threedvfinalcopy % *** Uncomment this line for the final submission

\definecolor{somegray}{rgb}{0.5, 0.5, 0.5}
\newcommand{\darkgrayed}[1]{\textcolor{somegray}{#1}}
\makeatletter
\newcommand*\titleheader[1]{\gdef\@titleheader{#1}}
\AtBeginDocument{%
  \let\st@red@title\@title
  \def\@title{%
    \vskip-3em
    \bgroup\normalfont\large\centering\@titleheader\par\egroup
    \vskip1.5em\st@red@title}
}
\makeatother

\titleheader{\darkgrayed{This paper has been accepted for publication at the \\
International Conference on 3D Vision (3DV), Prague, 2022.
\copyright IEEE}}

\title{Cross-Spectral Neural Radiance Fields}

\begin{document}

\author{Matteo Poggi$^*$ \hspace*{1cm} Pierluigi Zama Ramirez$^*$  \hspace*{1cm} Fabio Tosi$^*$ \\ Samuele Salti \hspace*{1cm} Stefano Mattoccia  \hspace*{1cm} Luigi Di Stefano \\
CVLAB, Department of Computer Science and Engineering (DISI)\\
University of Bologna, Italy\\
{\tt\small \{m.poggi, pierluigi.zama, fabio.tosi5\}@unibo.it} \\
{\tt\small Project page: \url{https://cvlab-unibo.github.io/xnerf-web}}
}

\twocolumn[{
\maketitle
\vspace{-1.2cm}
\begin{center}
    \captionsetup{type=figure}
    % our
    \includegraphics[width=0.7\textwidth]{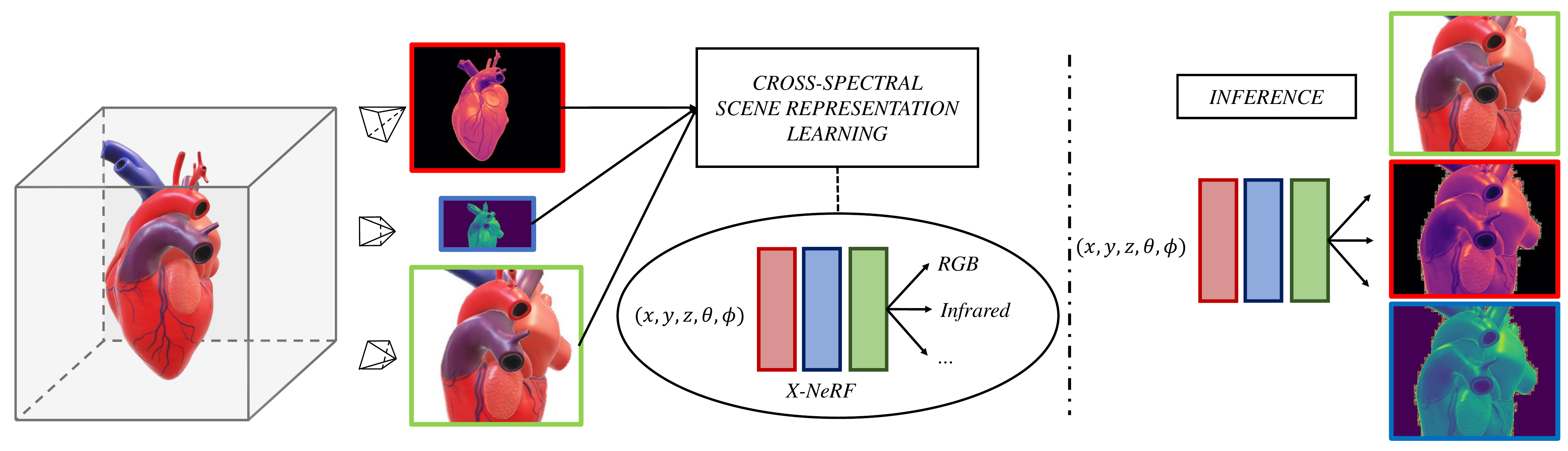}
    \vspace{-0.2cm}
    \captionof{figure}{\textbf{Cross-Spectral rendering with X-NeRF}. Given a set of images acquired from sensors with different light spectrum sensitivity -- such as infrared (red frame), RGB (green frame), multi-spectral (blue frame) -- resolution and field of view, we learn a shared cross-spectral scene representation, allowing for novel view synthesis across spectra.}
    \label{fig:teaser}
\end{center}
}]

\begin{abstract} We propose X-NeRF, a novel method to learn a Cross-Spectral scene representation given images captured from cameras with different light spectrum sensitivity, based on the Neural Radiance Fields formulation. X-NeRF optimizes camera poses across spectra during training and exploits Normalized Cross-Device Coordinates (NXDC) to render images of different modalities from arbitrary viewpoints, which are aligned and at the same resolution. Experiments on 16 forward-facing scenes, featuring color, multi-spectral and infrared images, confirm the effectiveness of X-NeRF at modeling Cross-Spectral scene representations.
\end{abstract}

\section{Introduction}
\blfootnote{$^*$ Joint first authorship.}Novel view synthesis, the task of synthesizing new images of an object or scene observed from arbitrary viewpoints, represents a long-standing problem at the intersection between vision and graphics. It enables several applications: video/image editing, virtual reality and so on.

In the last few years, a popular trend in novel view synthesis is to model scenes as \textit{implicit representations}. On this track,  Neural Radiance Fields (NeRF) \cite{Mildenhall2020ECCV} represents nowadays the most prominent paradigm to render images from arbitrary viewpoints, which yielded tremendous improvements in the quality of results. 
NeRF learns a scene representation as a 5D vector-valued function, modeled by a Multi-Layer-Perceptron (MLP), that outputs the emitted color $(R,G,B)$ and volume density $\sigma$ given as input a 3D location $(x,y,z)$ and a 2D viewing direction $(\theta, \phi)$.

However, we argue that representing a scene only through RGB colors may be limiting, as it fails to capture the richness of the spectral information around us. 
For instance, by capturing the surrounding visual information with only a single RGB camera, we cannot perceive natural phenomena that would require analysing the visible spectrum with a finer wavelength  granularity -- i.e. not only the classic red, green or blue channels -- or to go beyond  the visible range.
Such information could instead be gathered by sensors featuring different spectral sensitivity, such as multi-spectral (MS) or infrared (IR) cameras.  
Moreover, finding and analysing the correlations between spectra may help to gain a more in-depth understanding of natural processes. 
Consequently, to be able to reason on multi-spectral data, we would need to obtain a unified Cross-Spectral scene representation -- allowing for querying, for any single point, any of the information sensed across spectra. 

Based on the above observations, we propose for the first time a Cross-Spectral NeRF (\textbf{X-NeRF}), which can model scenes acquired from cameras featuring different spectral sensitivities.
Collecting images with a Cross-Spectral rig, we extend vanilla NeRF by training a single, shared network across spectra and learning a channel for each spectral band. 

However, this straightforward extension alone is not enough to properly model a unified, Cross-Spectral representation. Indeed, two main challenges arise when considering this peculiar setting. The first concerns the need for knowing exact camera poses for any of the images acquired from the different cameras and used to train NeRF. While this information can be obtained effortlessly when dealing with RGB images  \cite{schoenberger2016sfm}, it is not trivial to obtain camera poses according to a common reference system across the different image modalities. 
The second is linked to the marked differences between sensors -- resolution, field of view (FoV) -- which needs to be taken into account when casting rays across 3D space, to ensure that the very same point observed in the scene is reached by rays traced from the corresponding pixel in each camera. For instance, when processing forward-facing scenes, the standard Normalized Device Coordinates convention (NDC) \cite{Mildenhall2020ECCV,zhang2020nerfpp} fails at this.

Both the above challenges are addressed by our X-NeRF. 
As for the former, we obtain camera poses from RGB images \cite{schoenberger2016sfm} and let X-NeRF learns, during the training process, only the  relative poses of the other cameras -- which are supposed to be constant across views, since we assume cameras being rigidly mounted on a common rig -- to obtain the viewpoints for any modality starting from RGB images.
Concerning the latter,  we propose  Normalized Cross-Device Coordinates (NXDC) to align the ray sampling strategy across cameras, taking into account the different resolutions and FoVs so as to correctly map a single point perceived by any of the cameras to the very same 3D location.

As outcome, X-NeRF enables novel view synthesis across spectra and, more importantly, rendering of aligned spectral information from any viewpoint, as shown in \cref{fig:teaser}. We feel this latter aspect to be one major achievement of X-NeRF, since it yields the following  appealing results: i) during rendering, it realizes a \textit{virtual} Cross-Spectral camera, sensing a multitude of spectra from the very same viewpoint -- which does not occur when sensing the scene with the different cameras together, ii) given a real image acquired from a specific viewpoint, we can render the remaining modalities aligned to the real image itself, avoiding to address a non-trivial cross-modal matching problem \cite{zhi2018deep,tosi2022rgbms}, and iii) thanks to its continuous formulation, X-NeRF allows for super-solving low-resolution spectral data, e.g. so as to render MegaPixel MS data whereas existing MS cameras feature a dramatically lower resolution ($\sim$0.1Mp).

To evaluate the effectiveness of X-NeRF, we built a custom rig 
with a high-resolution RGB camera and two low-resolution IR and MS cameras, and used it to acquire a total of 16 forward-facing scenes with $\sim$30 different viewpoints for each modality, for a total of 90 views per scene, available in our project page. 
Our main contributions are:
\begin{itemize}
    \item[$\bullet$] We are the first to explore the problem of learning a Cross-Spectral scene representation using the Neural Radiance Field paradigm.
    
    \item[$\bullet$] To obtain camera poses, we learn the relative transformation between different sensors while training X-NeRF itself, thus avoiding non-trivial across spectra calibration/matching.

    \item[$\bullet$] We propose Normalized Cross-Device Coordinates (NXDC), to deal with coordinate system misalignment between modalities in forward-facing scenes.

    \item[$\bullet$] We propose a dataset of 16 forward-facing scenes acquired with sensors featuring three modalities (RGB, MS, and IR) used to train and validate our proposal.
    
\end{itemize}

\section{Related Work}

\textbf{Neural Radiance Fields.} 
Novel view synthesis has a rich history within the computer vision and computer graphics fields. Recent explicit methods based on deep learning train CNNs for this very purpose \cite{ZhouTFFS18,FlynnBDDFOST19,MildenhallSCKRN19,SrinivasanTBRNS19,LiXDS20,TuckerS20,LombardiSSSLS19,SitzmannTHNWZ19,HeCJS20}.  Nowadays, NeRF has become the dominant scene representation for view synthesis. It  allows for reconstructing photo-realistic novel views by means of a continuous volumetric fuction parameterized as a fully connected neural network, optimized by using a sparse set of input views. NeRF has inspired many subsequent works that extend its continuous neural volumetric representation in order to deal with different setups, e.g. dynamic scenes~\cite{Martin-BruallaR21,PumarolaCPM21,LiNSW21,XianHK021,GaoSKH21}, relighting~\cite{SrinivasanDZTMB21,ZhangSDDFB21,BossBJBLL21}, imperfect camera poses \cite{lin2021barf,wang2021nerf}, multi-resolution images \cite{Barron_2021_ICCV}, deformable agents~\cite{ParkSBBGSM21,TretschkTGZLT21,GafniTZN21,NoguchiSLH21,ParkSHBBGMS21} or to realize  generative models~\cite{SchwarzLN020,ChanMK0W21,KosiorekSZMSMR21}. Despite the impressive capability to represent realistic appearance, these works usually suffer from notable limitations such as 1) a long training process, 2) a slow rendering phase and 3) the requirement to perform a standalone training from scratch for any scene. This makes the aforementioned representations impractical for use in most applications that require real-time rendering.

\textbf{Faster NeRF Rendering.} 
Different approaches pursue speeding up of the volume rendering process run by MLP-based representations. Recent works combine a dense 3D grid of MLPs with empty space skipping and early termination \cite{ReiserPLG2021}, build and dynamically update an octree structure to avoid redundant MLP queries in free space \cite{LiuGLCT20}  or leverage explicit volumetric representations \cite{WizadwongsaPYS21,YuLTLNK2021,GarbinKJSV2021,HedmanSMBD2021}. Although the rendering speed up, gradient-based optimization cannot be used to directly optimize the data structures that are necessary for fast rendering. This means that a conversion step, from a trained model to the final representation that allows real-time rendering, is still needed.

\textbf{Faster NeRF Training.} 
Other recent works that focus on fewer input views bring faster convergence and, thus, a faster training process. Such methods typically rely on pre-training aimed at achieving generalization ~\cite{YuYTK21,WangWGSZBMSF21}, traditional Multi-View Stereo (MVS) approaches \cite{SRF,ChenXZZXYS21}, neural rays~\cite{LiuPLWWTZW21}, exploiting explicit representations~\cite{yu2021plenoxels} or combining them with implicit ones ~\cite{sun2021direct,mueller2022instant}.

\textbf{Multi-Spectral Imaging.} 
There exist several works in the field of multi-spectral (MS) imaging in the most diverse areas, ranging from robotics to automotive and from biometrics to surveillance. These applications  demand a combination of visible and non-visible wavelengths ranges such as Near infrared (NIR), short-wave infrared (SWIR) and mid-wave infrared (MWIR) . To name a few, some works adopt RGB-NIR for scene parsing \cite{choe2018ranus}  and recognition \cite{brown2011multi}  while others deploy NIR images for color enhancement \cite{zhang2008enhancing} and dehazing \cite{feng2013near}. Other sensors, such as thermal cameras, can directly measure long-wave infrared radiation of objects regardless of an external light source, and have deployed in pedestrian detection applications \cite{hwang2015multispectral,xu2017learning}. Moreover, cross-spectral matching represents another  challenging task that consists in recovering depth by finding correspondences between images with different spectra, in most cases by matching RGB-MS\cite{tosi2022rgbms},  RGB-IR\cite{chiu2011improving,mehltretter2018multimodal}, RGB-thermal\cite{pinggera122012cross} and RGB-NIR modalities \cite{zhi2018deep,shen2014multi,kim2016dasc,kim2016deep}.

\section{Method}
\label{sec:method}

In this section, we present our novel X-NeRF framework. We first introduce the NeRF background, then dig into the main novelties featured by X-NeRF.

\subsection{Background: Neural Radiance Field}
\label{sec:nerf}

Given an observed scene, NeRF \cite{Mildenhall2020ECCV} allows for novel view synthesis from arbitrary vantage points. This is achieved by training a neural network, i.e. a Multi Layer Perceptron (MLP), on a set of sparse images collected from different viewpoints. 
The MLP parametrises the \emph{Radiance Field} of the scene, i.e. a function of continuous 5D values $(x,y,z,\theta,\phi)$, where $\mathbf{x} = (x,y,z)$ are 3D coordinates in space and $(\theta,\phi)$ are viewing angles, function of camera pose $\Pi$. 
The direction can be also expressed as a 3D Cartesian unit vector $\mathbf{d}$.
Such a function produces a 4D output R,G,B,$\sigma$, encoding the color ($\mathbf{c}=(R,G,B)$) and volume density $(\sigma)$ of each 3D point in the scene. Specifically, the vanilla NeRF estimates color and density by means of two MLPs as $(\sigma, \mathbf{e}) = \text{MLP}^{\mathrm{(pos)}}(\mathbf{x})$, $\mathbf{c} = \text{MLP}^{\mathrm{(rgb)}}(\mathbf{e}, \mathbf{d})$, 
with $\sigma$ being interpreted as the differential probability of a ray terminating at $(x,y,z)$, and $\mathbf{e}$ being a feature embedding. 

\textbf{Volume Rendering.}\label{sec:volume_render} According to \cite{Max1995Optical}, the color $C(\mathbf{r})$ rendered from a camera ray $\mathbf{r}(t) = \mathbf{o} + t\mathbf{d}$  is obtained by solving the following integral: 
\begin{equation}
    C(\mathbf{r}) = \int_{t_n}^{t_f} T(t) \sigma(\mathbf{r}(t)) c(\mathbf{r}(t),\mathbf{d}) \textit{dt} 
\end{equation}
with $T(t)$ being the accumulated transmittance from $t_n$ to $t$ along ray $r$. The value of the integral is estimated via quadrature, by sampling $[t_n,t_f]$ in N evenly-spaced bins, with $t_n$ and $t_f$ being the near and far plane respectively.
\begin{equation}\label{eq:quadrature}
    C(\mathbf{r}) = \sum_{i=1}^{N} T_i(1-\text{exp}(-\sigma_i\delta_i))c_i, \hspace{0.4cm} T_i = \text{exp}\Big(-{\sum_{j=1}^{i-1}\sigma_j\delta_j}\Big)   
\end{equation}
with $\delta_i$ being the distance between adjacent samples $t_{i+1}$ and $t_i$. This procedure turns out to be equivalent to alpha compositing, assuming $\alpha_i = 1-\text{exp}(-\sigma_i\delta_i)$. Terms $T(i)\alpha_i$ act as a \textit{weight} $(w_i)$ for each point along the ray.

\textbf{Positional Encoding with Fourier Features.}
Traditionally, neural networks excel at learning low-frequency representations at the expense of high-frequency ones. As shown by NeRF \cite{Mildenhall2020ECCV}, encoding 3D coordinates $\mathbf{x}$ into a higher dimensional space allows to better recover the latter. This is achieved  by applying a so-called Fourier mapping $\gamma$ to each input component $p$ independently as $\gamma(p) = ( \sin{(2^0\pi p)}, \cos{(2^0\pi p)}, ... , \sin{(2^{L-1}\pi p)}, \cos{(2^{L-1}\pi p)})$.

\textbf{Ray Coordinates.}
According to the specific scene, two main coordinate systems are used to compute 3D coordinates of points laying along rays. In case of 360° scenes framing objects with masked backgrounds, conventional world coordinate systems are used, requiring the definition of near/far bounding planes. 
In case of forward-facing scenes, i.e. the camera rotation between views is small or absent, Normalized Device Coordinates (NDC) \cite{Mildenhall2020ECCV,zhang2020nerfpp} are used as standard convention, warping an 
infinitely deep camera frustum into a bounded $[-1,1]^3$ cube, where distance along the z-axis corresponds to disparity (inverse distance).
This parameterization optimizes the network capacity in a way that is consistent with the geometry of perspective projection, easing the problem itself -- in particular, in presence of large displacements between foreground and background \cite{zhang2020nerfpp}. 

\subsection{X-NeRF: Cross-Spectral NeRF}\label{sec:X-NeRF}

Differently from the original NeRF formulation, our goal is to obtain a Cross-Spectral neural scene representation.
We assume availability of $N_{m}$ cameras featuring different modalities (e.g., RGB, infrared, multi-spectral), mounted on a rigid system -- i.e. with fixed  relative poses between cameras.
Each camera with modality $m$ has a spatial resolution of $H_m \times W_m$ and $C_m$ number of channels, and it has been previously calibrated to estimate the intrinsic parameters $f_m^x, f_m^y, c_m^x, c_m^y$ (focal length and piercing point).
For each camera, we acquire a set of $N_{views}$ images from different viewpoints of the same scene. Thus, we gather a total of $N_{m} \times N_{views}$ images per scene.

We learn the Cross-Spectral scene representation as a function to map 5D coordinates $(x,y,z,\theta,\phi)$ into a volume density ($\sigma$) and $N_m$ output modalities, with each modality having $C_m$ channels, for a total of $\sum_{m}{C_m}$ channels. In our setup, we assume a shared volume density across modalities: this means rays emitted by the different sensors hit the very same 3D points and frame it in the images. This assumption would not hold in case this latter hypothesis is violated (e.g., when dealing with RGB and X-rays sensors).

At each optimization iteration, we select a training image with modality $m$, iterating over all the possible modalities at each step. Then, we sample a random batch of camera rays from the set of all its pixels.
For each ray $\mathbf{r}$, we then use the volume rendering procedure described in \cref{sec:volume_render} to estimate the response of that modality, $\hat{C}_m(\mathbf{r})$.
Our loss is simply the total squared error between the rendered and true pixel values for the considered modality:
\begin{equation}
    L_m = \sum_{\mathbf{r} \in R}{||\hat{C}_m(\mathbf{r}) - C_m(\mathbf{r})||^2_2}
    \label{eq:loss}
\end{equation}
where $R$ is the set of rays in each batch, and $C_m(\mathbf{r})$ and $\hat{C}_m(\mathbf{r})$ are the
ground truth and predicted modality for ray $\mathbf{r}$, respectively. 
Since each modality is acquired from different viewpoints, for a single ray $\mathbf{r}$ we can never compute the loss on multiple modalities. Thus, the training is carried out on the different modalities in interleaved manner.

\textbf{Pose Estimation.}
NeRF assumes camera poses to be known beforehand during training. 
This information is typically retrieved by means of COLMAP \cite{schoenberger2016sfm} on the set of RGB images  based on matching between image keypoints.
Since in our case we have images captured by cameras with different modalities, it is extremely hard to estimate reliable keypoints and descriptors amenable to perform matching across modalities.
A possible solution could be to apply COLMAP on each modality independently.
However, the estimated poses would be in different reference systems (typically the first frame of the sequence) and up to different scale factors, and would be not trivial to align all cameras in a shared reference system -- since it would require, again, matching across modalities.

However, as we assume cameras to be mounted on the same rigid rig, we can exploit COLMAP only on a single modality, and learn the relative poses between sensors as latent variables optimized during training, thereby avoiding the problem of matching pixel across spectra.
One may argue that relative poses could be estimated offline through calibration \cite{zhang2000flexible}. It is however non-obvious how to perform it across distant spectra, e.g. LWIR vs RGB may need ad-hoc calibration patterns \cite{shivakumar2020pst900}. More importantly, such poses would be metric and thus require alignment with the unknown COLMAP scale, a non-trivial problem itself.

Formally, given poses $\Pi^i_{m_{\alpha}}, i \in  \{1..N_{views} \}$ estimated by COLMAP on a reference modality $m_{\alpha}$ -- RGB in our setup -- we learn the relative poses $\Pi_{m_{\alpha} \rightarrow m}$ for any modality $m \neq m_{\alpha}$. These, multiplied by $\Pi^i_{m_{\alpha}}$, allow for obtaining poses $\Pi^i_{m}$ for any image collected by the camera of modality $m$. 
Relative poses are learned by back-propagating the loss from \cref{eq:loss} up to $\Pi_{m_{\alpha} \rightarrow m}$ as in \cite{wang2021nerf}

\begin{equation}
    \Pi^i_m = \Pi_{m_{\alpha} \rightarrow m} \times \Pi^i_{m_{\alpha}}, \quad\quad\quad \Pi^i_m = \text{arg}\min_{\Pi^i_m}L_m(\mathbf{r})
\end{equation}
We highlight that each pose $\Pi_{m_{\alpha} \rightarrow m}$ is learned based on the reconstruction loss of its modality $m$ solely, without enforcing any match across modalities. We will show empirically that this is sufficient to estimate consistent poses.

\begin{figure}[t]
    \centering
    \renewcommand{\arraystretch}{0.3}
    \renewcommand{\tabcolsep}{1pt}
    \begin{tabular}{ccccccccccc}
        \multicolumn{11}{c}{\includegraphics[width=0.45\textwidth]{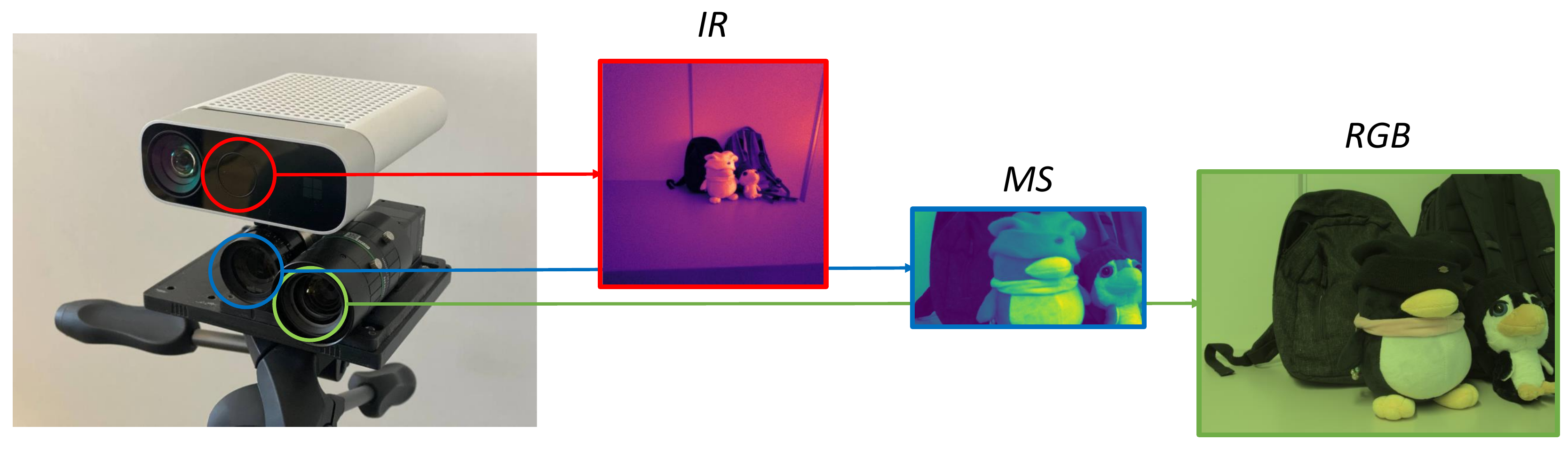}} \\
    \end{tabular}
    \vspace{-0.3cm}
    \caption{\textbf{Multi-modal camera rig.} We show the sensors suite used to collect our dataset.}
    \label{fig:rig_and_dataset}
    \vspace{-0.5cm}
\end{figure}

\textbf{NXDC: Normalized Cross Devices Coordinate.}
In forward-facing scenes, ray coordinates are usually expressed in Normalized Device Coordinates (NDC) \cite{Mildenhall2020ECCV}. 
However, NDC assumes that all images have been acquired by the same  camera. 
In our case, we have several cameras with marked differences such as resolution, focal length, etc.
Using every camera with its own intrinsics would cause a misalignment between ray coordinates across devices, leading X-NeRF to learn a non-registered scene representation -- i.e., the rendered images from the same point of view would be misaligned.
Thus, we introduce Normalized Cross-Device Coordinates (NXDC).
Assuming cameras looking in the $-z$ direction \footnote{\url{http://www.songho.ca/opengl/gl\_projectionmatrix.html}}, 3D points (in homogeneous coordinates) are projected according to perspective projection matrix $M$, function of near/far clipping planes $n,f$ and top/right scene bounds $r,t$ at near plane $n$

\begin{equation}
\pi(\mathbf{x}):
\begin{pmatrix}
\frac{nx}{r} \\
\frac{ny}{y} \\
\frac{-(f+n)z}{f-n}-\frac{2fn}{(f-n)z} \\
-z
\end{pmatrix}
\xrightarrow{}
\begin{pmatrix}
\frac{nx}{-rz}\\
\frac{ny}{-tz} \\
\frac{f+n}{f-n}+\frac{2fn}{z(f-n)}
\end{pmatrix}
\end{equation}

The projected point is now in NDC space, where the frustum has been mapped to a [-1,1]$^3$ cube.
Given a ray $\mathbf{o} +t\mathbf{d}$, we want to find the ray in NDC space that traces out the same point as the original ray (either at the same rate or not) -- i.e.,
compute the ray origin $\mathbf{o'}$ and direction $\mathbf{d'}$ such that, for every sampled point with $t$, there exists a $t'$ such that $\pi(\mathbf{o} + t \mathbf{d}) = \mathbf{o'} + t'\mathbf{d'}$.
We define:
\begin{equation}
       a_x = -\frac{n}{r} \hspace{0.3cm}
       a_y = \frac{-n}{t} \hspace{0.3cm}
       a_z = \frac{f+n}{f-n} \hspace{0.3cm}
       b_z = \frac{2fn}{f-n} \hspace{0.3cm}
\end{equation}
Assuming that the far scene bound is infinity, we obtain $a_z=1, b_z= 2n$.
If we consider the standard pinhole camera math, we can rewrite as:
\begin{equation}
    a_x = -\frac{f_x}{W/2} \hspace{1cm}
    a_y = -\frac{f_y}{H/2}
\end{equation}
where $f_x, f_y, H, W$ are the $x$ and $y$ focal lengths, height and width, respectively.

However, we employ different devices, thus we have different focal length, height and width for each camera. Simply using for each device its own parameters would lead to a different normalization across devices --i.e., different reference systems -- and thus X-NeRF will not be able to learn a unified scene representation with registered modalities. 
To overcome this problem, we constrain ratios $\frac{f_x}{W}$ and $\frac{f_y}{H}$ to be fixed across devices. To achieve this, we select the camera modality having minimum focal/image size ratio -- i.e., the largest FoV: 

\begin{equation}
    m_\beta^w \mid \frac{f^{m_\beta^w}_x}{W_{m_\beta^w}} = \min_m(\frac{f^m_x}{W_m}), \quad\quad 
    m_\beta^h \mid \frac{f^{m_\beta^h}_y}{H_{m_\beta^h}} = \min_m(\frac{f^m_y}{H_m})   
\end{equation}
Following NeRF derivation \footnote{\url{https://github.com/bmild/nerf/files/4451808/ndc\_derivation.pdf}}, we get $\mathbf{o'}$, $t'$, and $\mathbf{d'}$ according to fixed ratios as:
\begin{equation}
    \mathbf{o'} = 
    \begin{pmatrix}
        -\frac{m_\beta^w}{2} \frac{o_x}{o_z} \\
        -\frac{m_\beta^h}{2}  \frac{o_y}{o_z} \\
        1 + \frac{2n}{o_z} \\
    \end{pmatrix}
    \hspace{0.2cm}
    t' = \frac{td_z}{o_z + td_z}
    \hspace{0.2cm}
    \mathbf{d'} = 
    \begin{pmatrix}
        -\frac{m_\beta^w}{2} (\frac{d_x}{d_z} - \frac{o_x}{o_z}) \\
        \frac{m_\beta^h}{2}(\frac{d_y}{d_z} - \frac{o_y}{o_z}) \\
        -\frac{2n}{o_z}
    \end{pmatrix}
\end{equation}
In practice, this equals to padding images from sensors with lower FoV, while keeping focals unaltered. 
We dub the framework presented so far as \textbf{X-NeRF}.

\textbf{Speeding-up \textbf{X-NeRF}.} Finally, given the recent advances concerning fast training and rendering \cite{sun2021direct,yu2021plenoxels,mueller2022instant}, we also implement a faster variant of X-NeRF to bring it closer to unconstrained use in real applications. Specifically, we exploit a mixed implicit-explicit representation to speed up both phases.
Following DirectVoxGO (DVGO) \cite{sun2021direct}, we implement voxel grids -- actually, Multi-Plane Images (MPIs) in the case of forward-facing scenes \cite{sun2022improved} -- allowing for efficient queries in 3D space. Two structures, $\mathbf{M}^{\text{(dens)}}$ and $\mathbf{M}^{\text{(feat)}}$, are built respectively to encode density and feature embeddings, from which $\sigma$ is extracted by means of trilinear interpolation on the former, while color $\mathbf{c}$ is predicted by a shallow MLP queried with features interpolated from the latter as
$\sigma(\mathbf{x}) = \text{interp}(\mathbf{x}, \mathbf{M}^{\text{(dens)}})$, $\mathbf{c}(\mathbf{x},\mathbf{d}) = \text{MLP}^{\mathrm{(rgb)}} (\text{interp}(\mathbf{x}, \mathbf{M}^{\text{(feat)}}), \mathbf{x}, \mathbf{d})$.
Both are optimized through back-propagation during training. We dub this variant of our framework \textbf{X-DVGO}, since it extends the DVGO framework.

\section{Experimental Settings}
\subsection{Acquisition Setup and Dataset}
Our acquisition setup consists of three devices with different spectral sensitivity: a Ximea RGB camera equipped with a Sony IMX253LQR-C 12.4 Mpx sensor; an MS camera sensitive to 10 bands within the visible spectrum, based on an IM-SM4X4-VIS2 2.2 Mpx (one MS pixel capturing the 10 bands information uses a 4$\times$4 grid of native pixels, thus reducing the spatial resolution to 1/16); the passive infrared sensor of an Azure Kinect device, with a native resolution of 1Mpx.
Accordingly, our rig perceives 14 total channels across the sensed modalities.
The three cameras have been mounted on a rig, as depicted in \cref{fig:rig_and_dataset}, to acquire 16 indoor scenes -- since the Kinect IR sensor saturates outdoor -- with $\sim$30 images from each sensor per scene, 5 kept out for testing and the remaining used for training. As already mentioned, in this paper we focus on scenes acquired in forward-facing settings.
Examples of images acquired by our rig are shown in \cref{fig:rig_and_dataset}, where IR and MS images are encoded with colormaps \url{magma} and \url{viridis}, respectively (with MS being averaged over channels).

\subsection{Network Implementation and Training Details}

We implemented our framework using PyTorch. During training and testing, all images are normalized in $[0,1]$ over a single scene and modality, clipping intensities to the 99th percentile to filter intensity peaks in MS and IR images. Since IR images acquired by the Kinect are particularly noisy, we pre-process them by means of a $7\times7$ bilateral filter \cite{BILATERAL}. In the remainder of this sub-section we  report implementation details concerning X-NeRF and X-DVGO, whose output layers have been extended to predict multiple modalities as described in \cref{sec:X-NeRF}. 

\textbf{X-NeRF.} It is built on top of the NeRF-{}- codebase \cite{wang2021nerf}, 
which replicates the vanilla NeRF except for (i) not using hierarchical sampling strategy, (ii) reducing hidden layers dimension from 256 to 128 and (iii) sampling only 128 points along each ray, following \cite{wang2021nerf} to pursue computational efficiency. X-NeRF is trained for 5K epochs on each scene, i.e. $\sim$450k steps (150K per modality, $\sim$2.5 hours of overall training on a single 3090 RTX GPU).

\textbf{X-DVGO.} It is built on top of the DVGO codebase \cite{sun2021direct}, using 128 depth planes and a shallow MLP made of two hidden layers with 128 channels.
Following the default settings, X-DVGO is trained for 75K steps (25K per modality, taking about 15 minutes overall on a single 3090 RTX GPU), using a total variation regularizer \cite{rudin1994total} in addition to the rendering loss. Since we observed sub-optimal results when jointly optimizing camera poses with X-DVGO, leading to scarce alignment, we bootstrap it with camera poses learned after a few steps of X-NeRF training. However, training this latter for $<$10 minutes yields stable poses, ready for training X-DVGO.

\begin{table}[t]
    \centering
    \renewcommand{\tabcolsep}{2pt}
    \scalebox{0.8}{
    \begin{tabular}{;l|l|c|r;cc;}
    \hline
    \multicolumn{4}{;c;}{\textit{Configuration}} & \multicolumn{2}{c;}{\textit{Avg.}} \\
    \hline
    Model & Train & NXDC & Test & PSNR & SSIM \\
    \hline
    NeRF & RGB & - & RGB & \gold{32.44} & \gold{0.869} \\
    X-NeRF & RGB+MS & \ding{55} & RGB & \bronze{31.33} & \bronze{0.862} \\
    X-NeRF & RGB+MS & \checkmark & RGB & \silver{31.93} & \silver{0.864} \\
    \hline
    \hline
    NeRF & MS & - & MS & \silver{33.53} & \silver{0.917} \\
    X-NeRF & RGB+MS & \ding{55} & MS & \bronze{31.96} & \bronze{0.897} \\
    X-NeRF & RGB+MS & \checkmark & MS & \gold{33.87} & \gold{0.918} \\
    \hline
    \end{tabular}
    }
    \vspace{-0.3cm}
    \caption{\textbf{Bimodal Cross-Spectral rendering quality -- NeRF vs X-NeRF.} We report PSNR and SSIM averaged over the whole dataset. }
    \label{tab:rendering_bimodal}
    \vspace{-0.5cm}
\end{table}

\section{Experimental Results}

We evaluate X-NeRF on two main tasks: novel view synthesis and cross-modal image alignment.
In most tests, we highlight \gold{best}, \silver{second best} and \bronze{third best} methods according to average performance. Results on single scenes are reported in the \textbf{supplementary material}.

\subsection{Novel View Synthesis}

We start by evaluating the quality of novel views rendered by X-NeRF and X-DVGO for any modality. Specifically, given a single modality -- MS, for instance -- we render images from the viewpoints of that camera alone -- e.g., the MS camera -- and measure the quality over such modality -- e.g., MS predictions by the MLP.
To this aim, we report the PSNR and SSIM metrics \cite{Mildenhall2020ECCV} (the higher, the better), leaving out LPIPS -- since meaningful for RGB images only.

\begin{figure*}[t]
    \centering
    \renewcommand{\arraystretch}{0.3}
    \begin{tabular}{lrclrclrclr}
    \multicolumn{2}{c}{\includegraphics[trim={4cm 1cm 3cm 1cm},clip,height=0.16\textwidth]{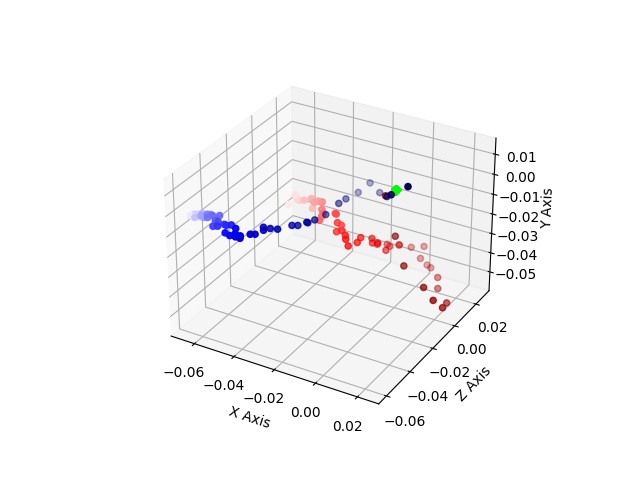}} & \quad &
    \multicolumn{2}{c}{\includegraphics[trim={4cm 1cm 3cm 1cm},clip,height=0.16\textwidth]{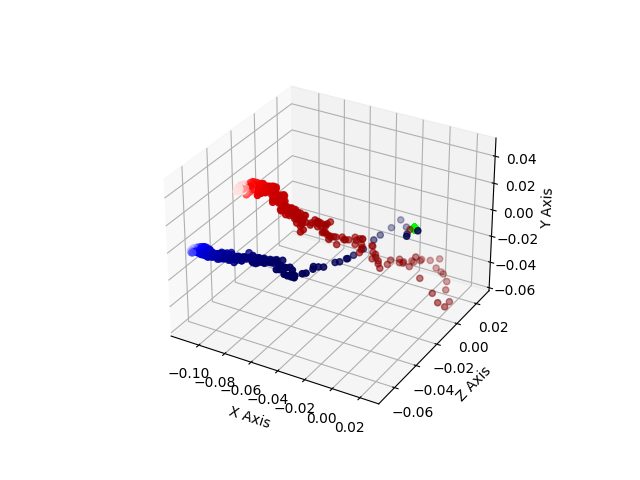}} & \quad &
    \multicolumn{2}{c}{\includegraphics[trim={4cm 1cm 3cm 1cm},clip,height=0.16\textwidth]{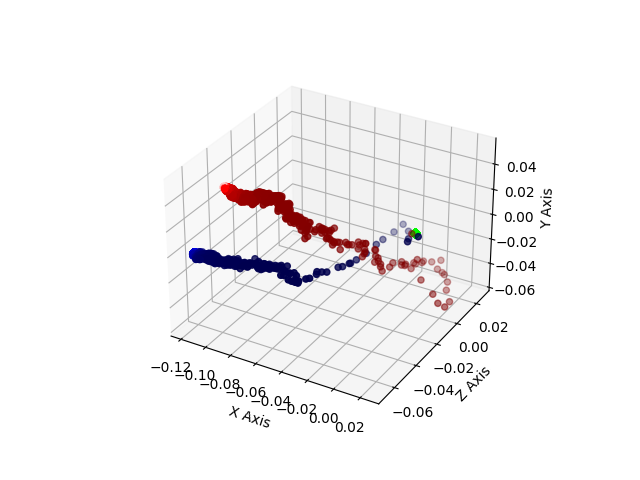}} & \quad &
    \multicolumn{2}{c}{\includegraphics[height=0.16\textwidth]{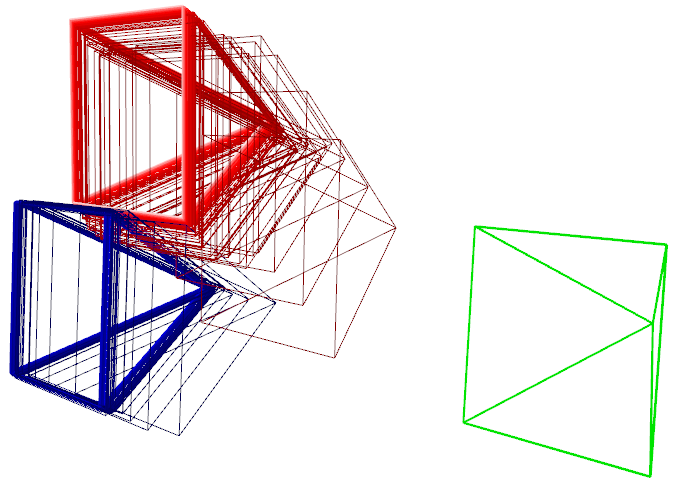}}
    \\ 
    \multicolumn{2}{c}{\includegraphics[width=0.16\textwidth]{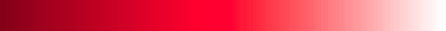}} & \quad &
    \multicolumn{2}{c}{\includegraphics[width=0.16\textwidth]{images/poses/seismic_r.png}} & \quad &
    \multicolumn{2}{c}{\includegraphics[width=0.16\textwidth]{images/poses/seismic_r.png}} \\
    \multicolumn{2}{c}{\includegraphics[width=0.16\textwidth]{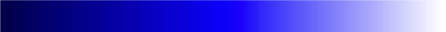}} & \quad &
    \multicolumn{2}{c}{\includegraphics[width=0.16\textwidth]{images/poses/seismic.png}} & \quad &
    \multicolumn{2}{c}{\includegraphics[width=0.16\textwidth]{images/poses/seismic.png}} \\
    \tiny 0 & \tiny 50 & \quad & \tiny 0 & \tiny 500 & \quad & \tiny 0 & \tiny 5000 \\
    \end{tabular}
    \vspace{-0.3cm}
    \caption{\textbf{Relative cameras positions during training.} Left: we show how camera centers translate in 3D space in 50, 500 and 5000 epochs. Right: we display camera frustums. After $\sim$250 epochs ($\sim$7.5 minutes), cameras become stable.}
    \label{fig:poses_trend}
    \vspace{-0.35cm}
\end{figure*}

\textbf{Bimodal Cross-Spectral Radiance Field.} As a first experiment, we train X-NeRF to deal with images belonging to two modalities, RGB and MS, and compare it with a vanilla NeRF trained on single modalities alone. In this case, a single relative pose between RGB and MS camera is learned during optimization.
\cref{tab:rendering_bimodal} collects the outcome of this evaluation. Each row corresponds to a specific model (NeRF or X-NeRF), the modalities used for training, optional use of NXDC space and the testing modality -- which also bounds rendering resolution.

Considering RGB rendering, we can notice that the vanilla NeRF trained on RGB images alone achieves, overall, the best performance. This is not surprising, since the additional MS images processed by X-NeRF are at much lower resolution (about 100$\times$ smaller) and, of course, of different modality. However, the drop is moderate thanks to the MS bands partially overlapping the RGB ones. Moreover, we can appreciate how the drop is smaller when the NXDC space is used, showing how our proposal  favors learning a joint representation of the two modalities by the MLP. 
By looking at MS rendered images, we can appreciate how X-NeRF outperforms vanilla NeRF with NXDC, rendering higher-quality images. The higher-resolution of RGB images and the partial overlap with MS ones allows for such improvement, while X-NeRF using NDC cannot exploit such advantage.
Further experiments concerning NDC and XNDC are reported as \textbf{supplementary material}.

To summarize, X-NeRF achieve comparable performance with respect to NeRF when rendering RGB images -- thanks to the NXDC space -- while it can effectively exploit the learned cross-spectral representation of the scene to improve the quality of rendered MS images.

\begin{table}[t]
    \centering
    \renewcommand{\tabcolsep}{2pt}
    \scalebox{0.8}{
    \begin{tabular}{;l|l|c|r;cc;}
    \hline
    \multicolumn{4}{;c;}{\textit{Configuration}} & \multicolumn{2}{c;}{\textit{Avg.}}  \\
    \hline
    Model & Train & NXDC & Test & PSNR & SSIM \\
    \hline
    NeRF & RGB & - & RGB & \gold{32.44} & \gold{0.869}\\
    X-NeRF & RGB+MS+IR & \ding{55} & RGB & \bronze{30.43} & \bronze{0.856}\\
    X-NeRF & RGB+MS+IR & \checkmark & RGB & \silver{31.61} & \silver{0.862}\\
    \hline
    \hline
    NeRF & MS & - & MS & \gold{33.53} & \gold{0.917}\\
    X-NeRF & RGB+MS+IR & \ding{55} & MS & \bronze{30.87} & \bronze{0.870}\\
    X-NeRF & RGB+MS+IR & \checkmark & MS & \gold{33.53} & \silver{0.914}\\
    \hline
    \hline
    NeRF & IR & - & IR & \gold{33.26} & \gold{0.897}\\
    X-NeRF & RGB+MS+IR & \ding{55} & IR & \bronze{31.60} & \bronze{0.869}\\
    X-NeRF & RGB+MS+IR & \checkmark & IR & \silver{32.44} & \silver{0.879}\\
    \hline
    \end{tabular}
    }
    \vspace{-0.3cm}
    \caption{\textbf{Trimodal Cross-Spectral rendering quality -- NeRF vs X-NeRF.} We report PSNR and SSIM averaged over the whole dataset. }
    \label{tab:rendering_trimodal}
    %\vspace{-0.75cm}
\end{table}

\begin{table}[t]
    \centering
    \renewcommand{\tabcolsep}{2pt}
    \scalebox{0.8}{
    \begin{tabular}{;l|l|c|r;cc;}
    \hline
    \multicolumn{4}{;c;}{\textit{Configuration}} & \multicolumn{2}{c;}{\textit{Avg.}} \\
    \hline
    Model & Train Time & NXDC & Test & PSNR & SSIM \\
    \hline
    X-NeRF & $\sim$2.5 hours & \checkmark & RGB & \silver{31.61} & \silver{0.862}\\
    X-DVGO & $\sim$22.5 mins & \checkmark & RGB & \gold{31.77} & \gold{0.887}\\
    \hline
    \hline
    X-NeRF & $\sim$2.5 hours & \checkmark & MS & \gold{33.53} & \silver{0.914} \\
    X-DVGO & $\sim$22.5 mins & \checkmark & MS & \silver{33.22} & \gold{0.922}\\
    \hline
    \hline
    X-NeRF & $\sim$2.5 hours & \checkmark & IR & \gold{32.44} & \silver{0.879}\\
    X-DVGO & $\sim$22.5 mins & \checkmark & IR & \silver{31.60} & \gold{0.908}\\
    \hline
    \end{tabular}
    }
    \vspace{-0.3cm}
    \caption{\textbf{Trimodal Cross-Spectral rendering quality -- X-NeRF vs X-DVGO.} We report PSNR and SSIM averaged over the whole dataset.}
    \label{tab:rendering_dvgo}
    \vspace{-0.5cm}
\end{table}

\textbf{Trimodal Cross-Spectral Radiance Field.} We now add a further modality to X-NeRF, training it to render jointly RGB, MS and IR images. With this setup we will conduct all the following experiments.
\cref{tab:rendering_trimodal} collects the outcome of this experiment. We can notice how, on any modality, the vanilla NeRF trained on the single modality alone achieves the best results on average. Again, we feel this trend to be not surprising, given the variety of content framed by the three modalities and the very different resolutions of each. However, we can observe once again how NXDC  allows for the smallest drops. 
By looking at the individual modalities, on average X-NeRF achieves equivalent performance with respect to vanilla NeRF on MS images, while dropping on RGB and IR rendered images.

\begin{table*}[t]
    \centering
    \renewcommand{\tabcolsep}{8pt}
    \scalebox{0.7}{
    \begin{tabular}{;l|r;lcc|lcc|lcc|lcc;}
    \hline
    \multicolumn{2}{;c;}{\textit{Configuration}} & \multicolumn{3}{c|}{\textit{$\sim$16.5 mins}} & \multicolumn{3}{c|}{\textit{$\sim$22.5 mins}} & 
    \multicolumn{3}{c|}{\textit{$\sim$30 mins}} &
    \multicolumn{3}{c;}{\textit{$\sim$1 hour}} \\ 
    \hline
    & & Pose & & & Pose & & & Pose & & & Pose & & \\ 
    Model & Test & epochs & PSNR & SSIM & epochs & PSNR & SSIM & epochs & PSNR & SSIM & epochs & PSNR & SSIM \\ 
    \hline
    X-NeRF & RGB & 550 & 28.51 & 0.849 & 750 & 29.18 & 0.852 & 1000 & 29.67 & 0.854 & 2000 & 30.65 & 0.858 \\
    X-DVGO & RGB & 50 & 31.55 & 0.887 & 250 & 31.77 & 0.887 & 500 & 31.83 & 0.888 & 1500 & 31.81 & 0.887 \\ 
    \hline
    \hline
    X-NeRF & MS & 550 & 28.49 & 0.850 & 750 & 29.50 & 0.864 & 1000 & 30.38 & 0.877 & 2000 & 32.19 & 0.901 \\
    X-DVGO & MS & 50 & 32.92 & 0.918 & 250 & 33.22 & 0.922 & 500 & 33.28 & 0.923 & 1500 & 33.37 & 0.923 \\ 
    \hline
    \hline
    X-NeRF & IR & 550 & 28.55 & 0.838 & 750 & 29.55 & 0.849 & 1000 & 30.38 & 0.877 & 2000 & 31.30 & 0.869 \\
    X-DVGO & IR & 50 & 31.40 & 0.906 & 250 & 31.60 & 0.908 & 500 & 31.57 & 0.908 & 1500 & 31.62 & 0.909 \\ 
    \hline
    \end{tabular}
    }
    \vspace{-0.3cm}
    \caption{\textbf{Trimodal Cross-Spectral rendering quality - fixed time budget.} We report PSNR and SSIM (dataset average) under different training schedules.}
    \label{tab:rendering_dvgo_over_time}
    \vspace{-0.5cm}
\end{table*}

\textbf{Learned Poses Analysis.} We now inquire about how relative poses between RGB-MS and RGB-IR cameras are optimized by X-NeRF during training. \cref{fig:poses_trend} shows how MS and IR cameras centers move during training, according to the relative pose learned with respect to the RGB camera (green) after a certain number of epochs, encoded in blue (MS) and red (IR) color intensities respectively. According to colors being normalized over different epoch ranges, specifically 50, 500 and 5000, we can notice how after roughly 250 epochs the relative poses get stable and very close to those obtained after an entire training cycle. We can notice how the camera visualized in \cref{fig:poses_trend} (right) are placed as in the real rig shown in \cref{fig:rig_and_dataset}.

\textbf{Speeding-up Cross-Spectral Radiance Fields.} We now evaluate the rendering performance of the X-NeRF accelerated variant, namely X-DVGO. To train X-DVGO on a single scene, we bootstrap camera poses by training for 250 epochs X-NeRF on the same scene -- taking about 7.5 minutes. Then, we freeze relative poses and start training X-DVGO.
\cref{tab:rendering_dvgo} shows a comparison between X-NeRF and X-DVGO, trained on RGB, MS and IR modalities jointly and tested to render any of the three. In general, when rendering MS and IR images the two achieve very similar performance on average, with X-DVGO achieving slightly lower PSNR and higher SSIM scores, while when rendering RGB images X-DVGO outperforms X-NeRF on both metrics while training approximately 8$\times$ times faster (i.e., 7.5 plus 15 minutes versus 2.5 hours).

We now study the impact of the bootstrapped poses on X-DVGO performance, aimed at assessing the advantages it yields in terms of time required for training. In \cref{tab:rendering_dvgo_over_time} we report rendering performance by X-DVGO when trained with poses being optimized for different amounts of epochs. We compare it to X-NeRF trained for an amount of time equal to the total time required by X-DGVO (i.e., bootstrapping plus actual 25K steps of training). We can notice how poses optimized for 50 epochs only already yields render quality not that far from those by X-NeRF trained for an entire cycle (5K epochs), with only 16.5 minutes of total training. With the very same time budget, X-NeRF achieves remarkably worse results.
Some improvements are achieved by X-DVGO using poses optimized for 250 epochs, while elongating the poses initialization process for more epochs does not allow for further significant improvements. X-NeRF still results inferior in rendering quality when limiting the time budget up to one hour, confirming that X-DVGO achieves a better trade-off in terms of training time/rendering quality.

\begin{table}[t]
    \centering
    \renewcommand{\tabcolsep}{2pt}
    \scalebox{0.8}{
    \begin{tabular}{;l|l|c;ccc;}
    \hline
    \multicolumn{3}{;c;}{\textit{Configuration}} & \multicolumn{3}{c;}{\textit{Avg.}} \\
    \hline
    & & & RGB & MS & IR \\
    Model & Train Time & NXDC & 1694$\times$3434 & 254$\times$510 & 181$\times$363 \\
    \hline
    X-NeRF & $\sim$ 2.5 hours & \ding{55} & \bronze{0.214} & \bronze{0.234} & \bronze{0.277} \\
    X-NeRF & $\sim$ 2.5 hours & \checkmark & \gold{0.668} & \gold{0.667} & \gold{0.672}\\
    X-DVGO & $\sim$ 22.5 mins & \checkmark & \silver{0.632} & \silver{0.630} & \silver{0.639}\\ 
    \hline
    \end{tabular} 
    }
    \vspace{-0.2cm}
    \caption{\textbf{Cross-Spectral alignment quality.} We report MI averaged over the whole dataset. }
    \label{tab:rendering_registration}
    \vspace{-0.3cm}
\end{table}

\begin{figure*}[!ht]
    \centering
    \renewcommand{\arraystretch}{0.3}
    \renewcommand{\tabcolsep}{1pt}
    \begin{tabular}{cccccccccc}
        & \tiny \textit{IR} & \tiny \textit{RGB} & \tiny \textit{MS} & \tiny \textit{Depth} & \quad & \tiny \textit{IR} & \tiny \textit{RGB} & \tiny \textit{MS} & \tiny \textit{Depth} \\
        \rotatebox[origin=l]{90}{\tiny Real images} & \includegraphics[width=0.12\textwidth]{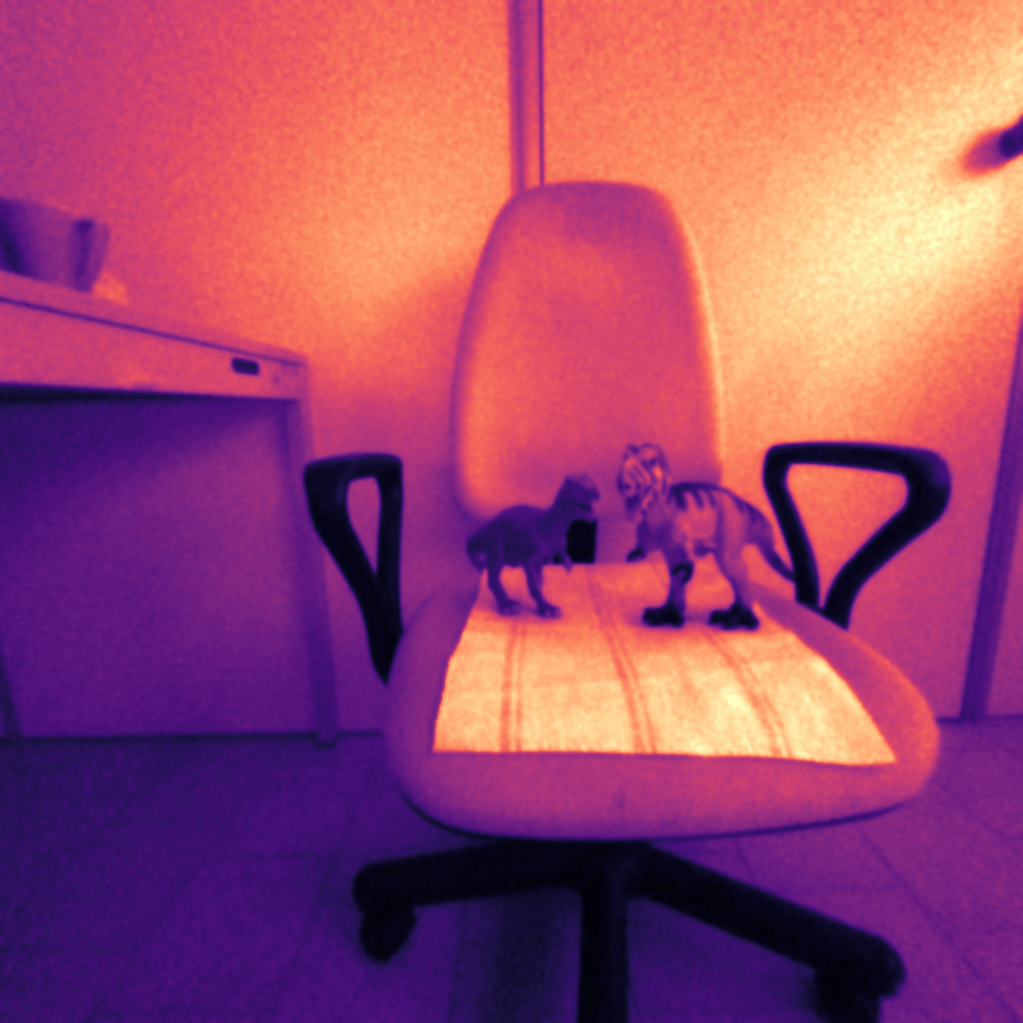} &
        \includegraphics[width=0.12\textwidth]{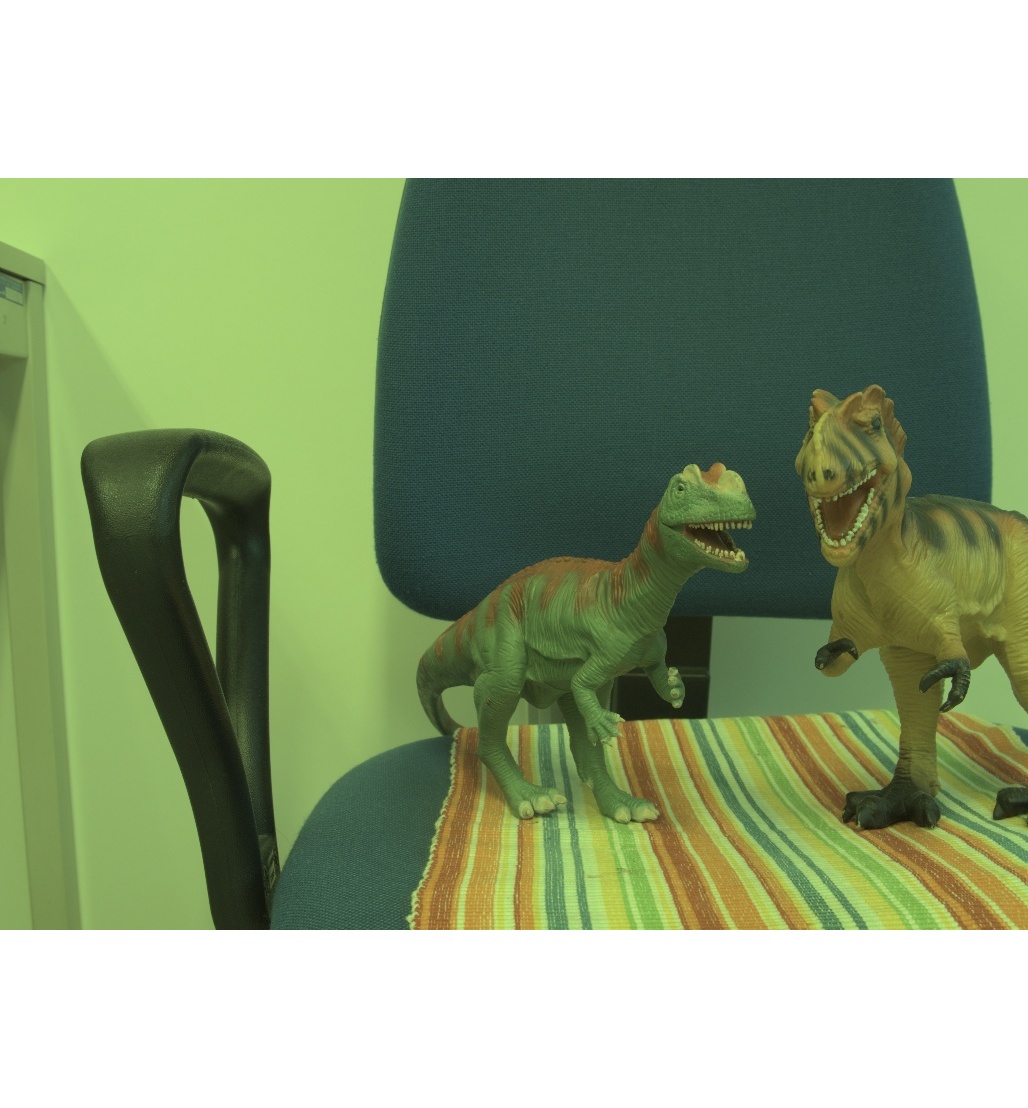} &
        \includegraphics[width=0.12\textwidth]{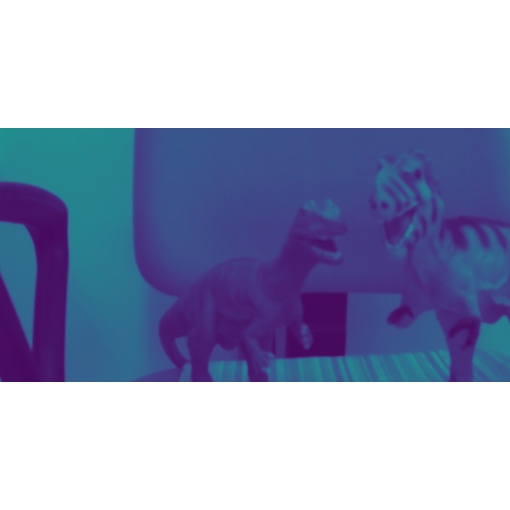} & & \quad & 
        \includegraphics[width=0.12\textwidth]{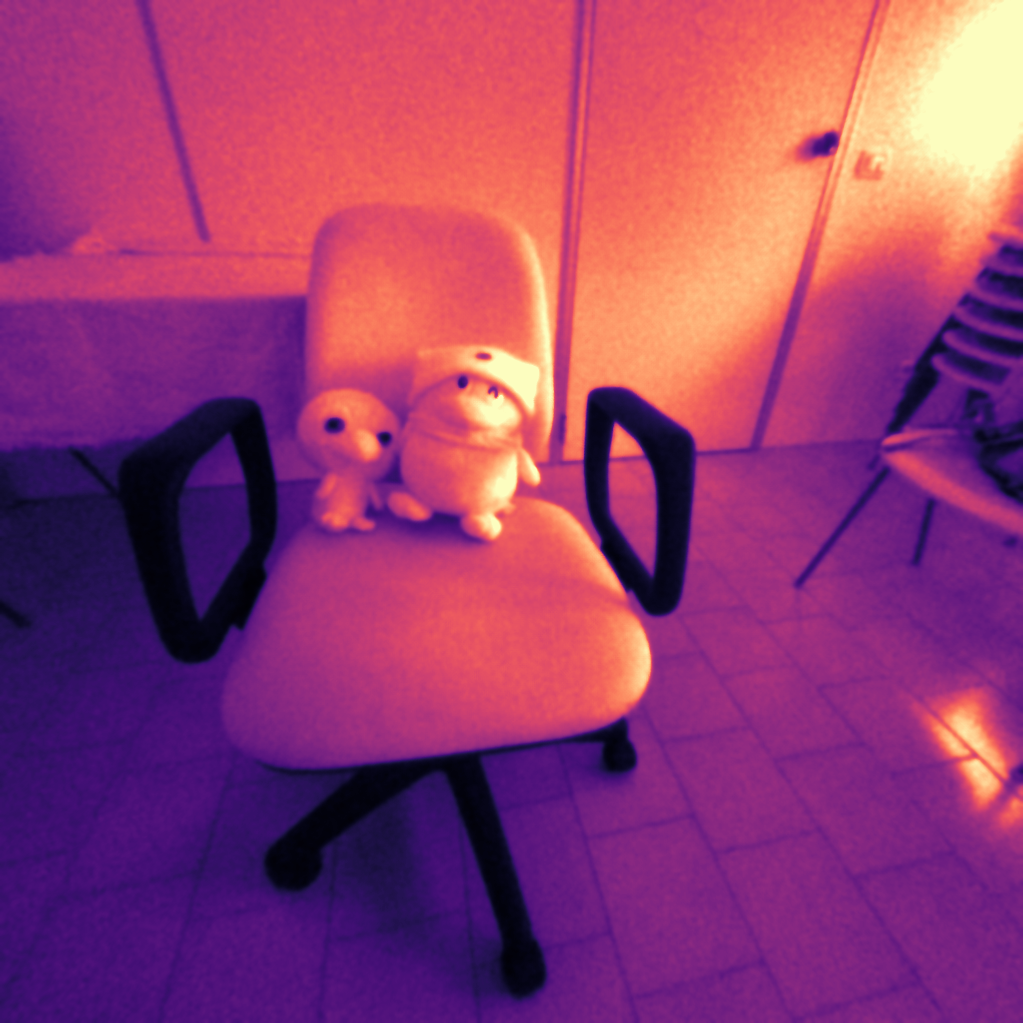} &
        \includegraphics[width=0.12\textwidth]{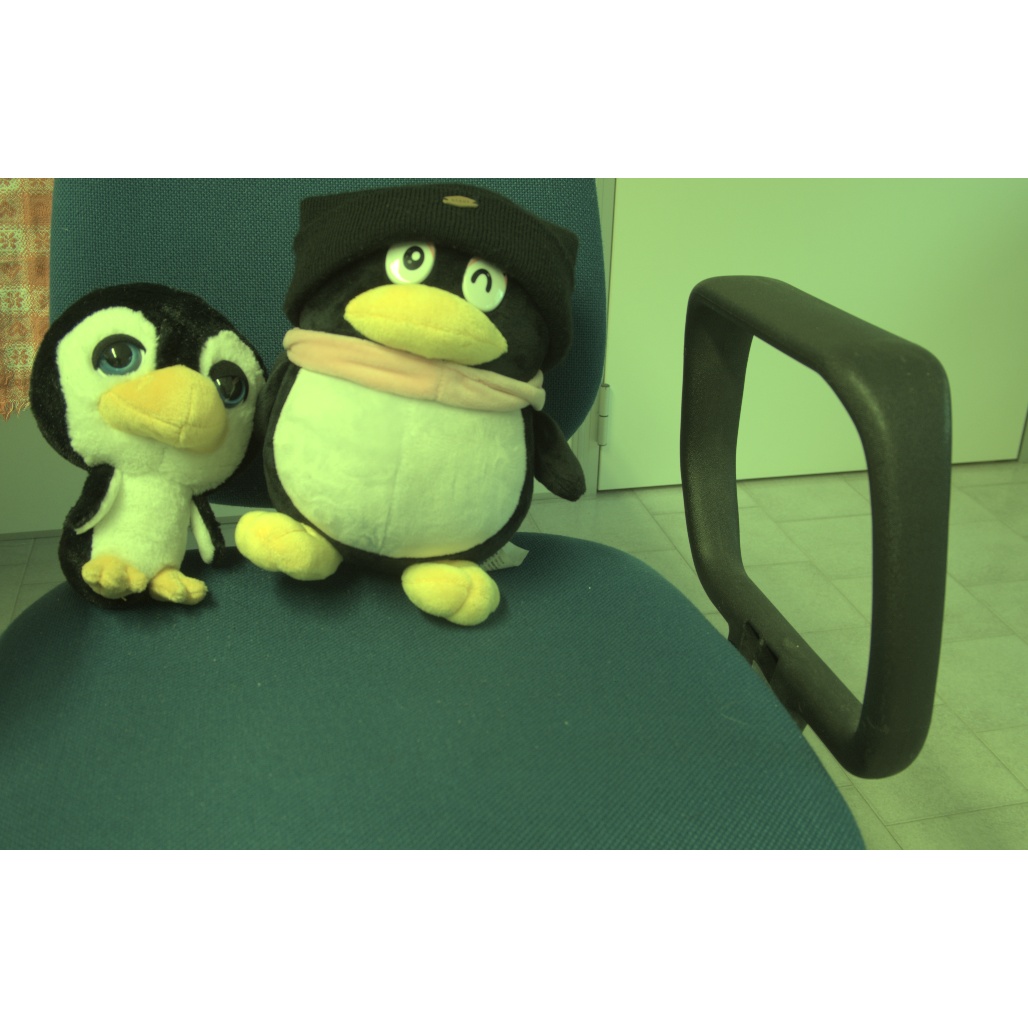} &
        \includegraphics[width=0.12\textwidth]{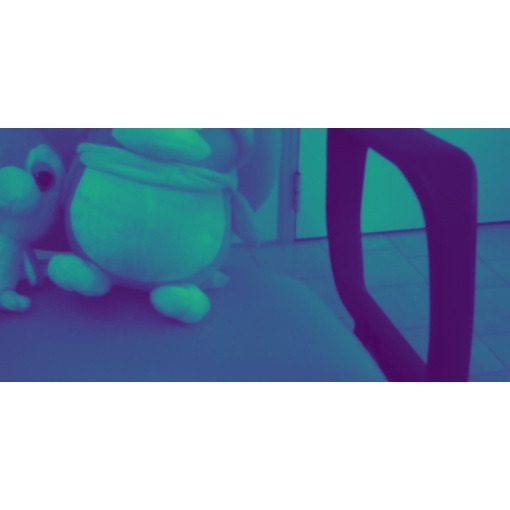}
        \\
        \rotatebox[origin=l]{90}{\tiny NDC} & \includegraphics[width=0.12\textwidth]{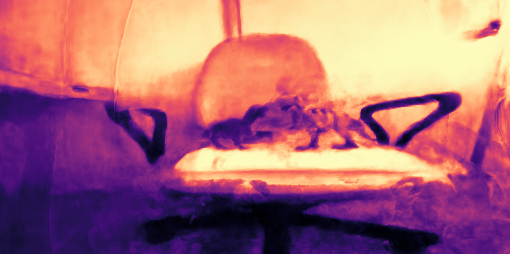} &
        \includegraphics[width=0.12\textwidth]{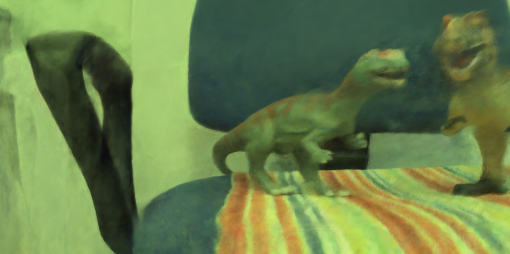} &
        \includegraphics[width=0.12\textwidth]{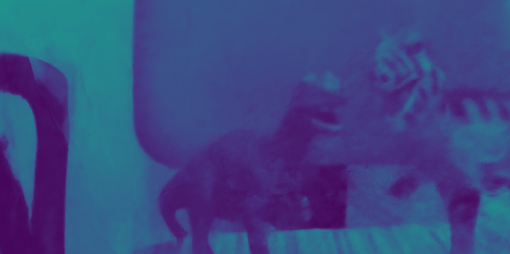} &
        \includegraphics[width=0.12\textwidth]{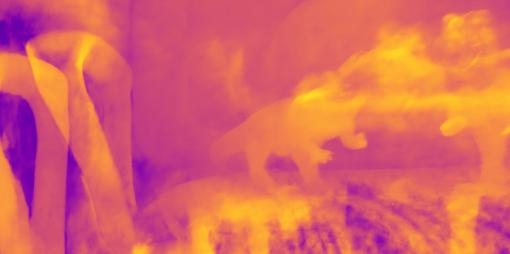}
        & \quad &
        \includegraphics[width=0.12\textwidth]{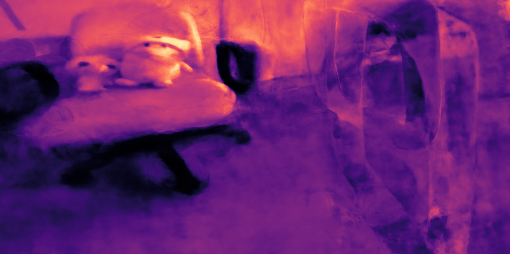} &
        \includegraphics[width=0.12\textwidth]{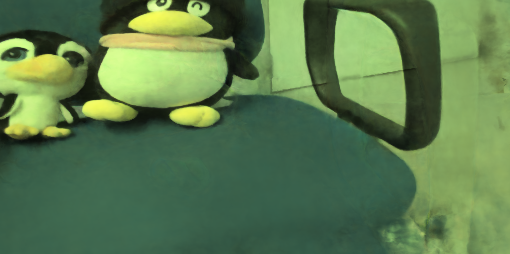} &
        \includegraphics[width=0.12\textwidth]{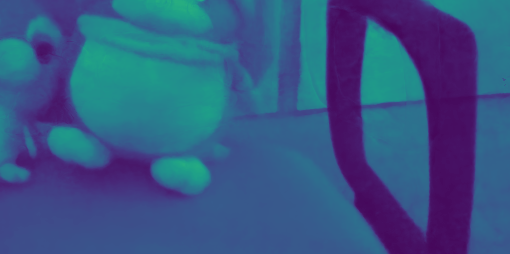} &
        \includegraphics[width=0.12\textwidth]{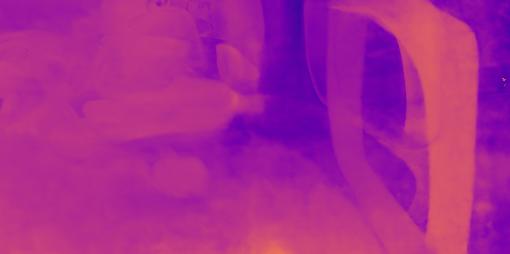}
        \\
        
        \rotatebox[origin=l]{90}{\tiny NXDC} &\includegraphics[width=0.12\textwidth]{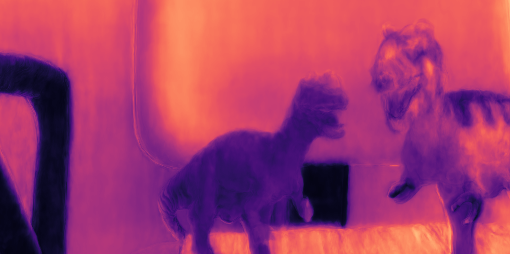} &
        \includegraphics[width=0.12\textwidth]{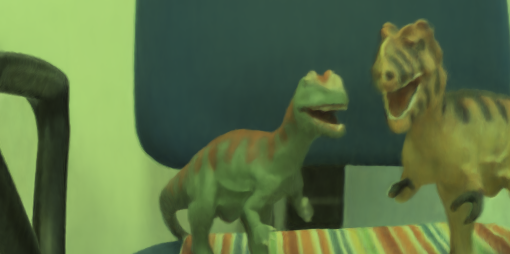} &
        \includegraphics[width=0.12\textwidth]{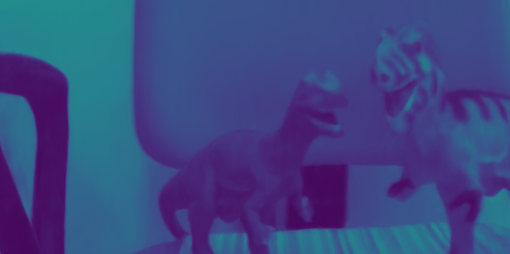} &
        \includegraphics[width=0.12\textwidth]{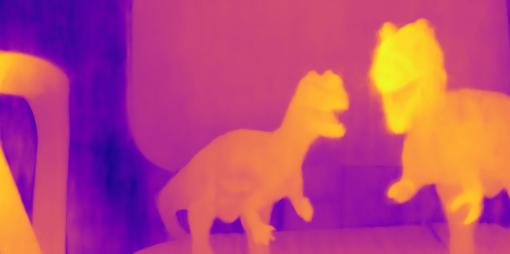} 
        & \quad &
        \includegraphics[width=0.12\textwidth]{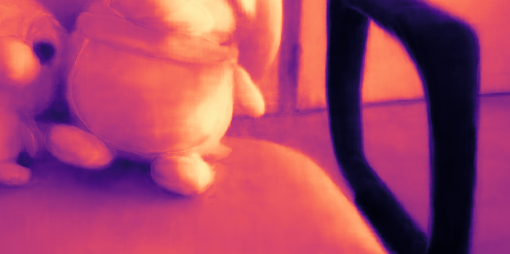} &
        \includegraphics[width=0.12\textwidth]{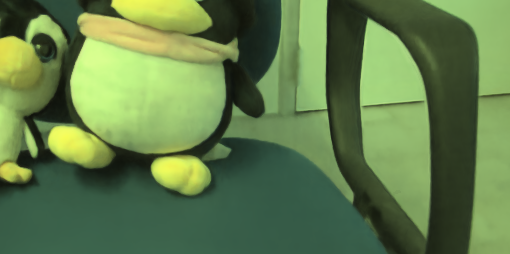} &
        \includegraphics[width=0.12\textwidth]{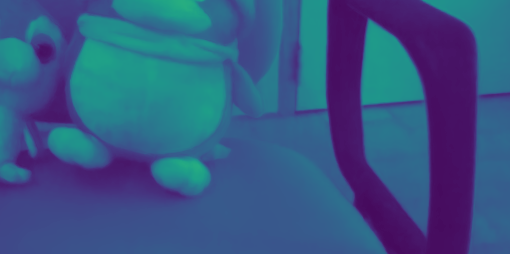} &
        \includegraphics[width=0.12\textwidth]{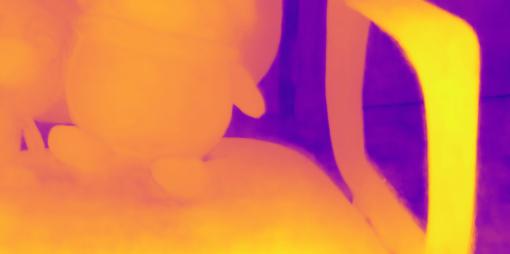}\\
        & \multicolumn{4}{c}{\tiny \textit{Dino}} &  \quad & \multicolumn{4}{c}{\tiny \textit{Penguin}} \\
    \end{tabular}
    \vspace{-0.3cm}
    \caption{\textbf{Cross-spectral rendering, qualitative examples.} Top: real images collected with our rig, followed by images and depth maps rendered by X-NeRF using NDC (middle) or NXDC (bottom), both assuming the MS camera viewpoint.}
    \label{fig:aligned_rendering}
    \vspace{-0.3cm}
\end{figure*}

\begin{table*}[!ht]
    \centering
    \renewcommand{\tabcolsep}{4pt}
    \scalebox{0.75}{
    \begin{tabular}{;l;ccc|ccc|ccc|ccc|}
    \hline
    \multicolumn{1}{;c;}{\textit{Configuration}} & \multicolumn{3}{c|}{\textit{$\sim$16.5 mins}} & \multicolumn{3}{c|}{\textit{$\sim$22.5 mins}} & 
    \multicolumn{3}{c|}{\textit{$\sim$30 mins}} &
    \multicolumn{3}{c|}{\textit{$\sim$1 hour}} \\
    \hline
    & RGB & MS & IR & RGB & MS & IR & RGB & MS & IR & RGB & MS & IR \\
    Model & 1694$\times$3434 & 254$\times$510 & 181$\times$363 & 1694$\times$3434 & 254$\times$510 & 181$\times$363 & 1694$\times$3434 & 254$\times$510 & 181$\times$363 & 1694$\times$3434 & 254$\times$510 & 181$\times$363 \\
    \hline
    X-NeRF & 0.718 & 0.714 & 0.720 & 0.698 & 0.700 & 0.703 & 0.694 & 0.694 & 0.701 & 0.681 & 0.681 & 0.686\\
    X-DVGO & 0.577 & 0.573 & 0.586 & 0.632 & 0.630 & 0.639 & 0.639 & 0.636 & 0.646 & 0.650 & 0.649 & 0.656 \\
    \hline
    \end{tabular} 
    }
    \vspace{-0.3cm}
    \caption{\textbf{Cross-Spectral alignment -- fixed time budget.} We report MI (dataset average) under different training schedules. }
    \label{tab:rendering_registration_over_time}
    \vspace{-0.5cm}
\end{table*}

\subsection{Cross-modal Alignment}

To conclude, we assess how effective X-NeRF and X-DVGO are at rendering images aligned across spectra, i.e. so as to create a \textit{virtual} Cross-Spectral camera.
This evaluation is carried out by rendering images according to each of the three cameras viewpoints, and thus at the three different resolutions they are characterized by, which are then cropped to match the area common to the three -- i.e., the one observed by the camera with the narrowest FoV, the MS camera in our case. This results in evaluating on 1694$\times$3434 images when rendering from RGB camera viewpoints, 254$\times$510 from MS viewpoints and 181$\times$363 from the IR cameras.
We compute pair-wise Mutual Information (MI, the higher the better) \cite{MUTUAL_INFORMATION} across the three modality pairs, and then average the three scores we obtain.

\cref{tab:rendering_registration} collects the outcome of this evaluation, involving X-NeRF -- without and with NXDC  -- and X-DVGO. We can notice how the MI across modalities is very low when X-NeRF uses the classical NDC: indeed, the MLP learns to render the three different modalities by casting rays in very different regions of the 3D space, resulting in unaligned rendered images. On the contrary, NXDC allows for learning much better aligned representations, thus achieving much higher MI scores. 
We can observe this effect also qualitatively, by looking at images and depth maps rendered by X-NeRF. \cref{fig:aligned_rendering} reports two samples from the \textit{Dino} and \textit{Penguin} scenes of our collected dataset, followed by images rendered from MS viewpoint by X-NeRF, trained with NDC or NXDC convention. We can notice how the latter allows for rendering images that are aligned across spectra, and properly models the 3D space as we can notice by observing the rendered depth maps. 

X-DVGO achieves results almost equivalent to X-NeRF, resulting in slightly lower MI scores. In \cref{tab:rendering_registration_over_time}, we show average MI scores at each resolution achieved when bootstrapping X-DVGO with poses initialized for different amounts of epochs, i.e. the same reported in \cref{tab:rendering_dvgo_over_time}. As we observed for rendering results, after 250 epochs the improvement almost saturates. By training X-NeRF with the same time budgets, alignment slightly reduces over time to favor the rendering quality over single modalities. The \textbf{supplementary material} provides more qualitative results.

\subsection{Failure Cases and Limitations}

Despite the high quality of both rendering and alignment yielded by X-NeRF, the task we are facing and the hypotheses under which we operate -- partially known camera poses and very different sensors modalities, resolutions, focals and FoVs -- are very challenging, thus some failure cases occur. Specifically, for some scenes X-NeRF gets stuck into local minima and cannot align the three modalities at their best. We show in the \textbf{supplementary material} some examples of this occurrences, with two modalities being properly aligned and the third one resulting slightly drifted.

\section{Conclusion}
We proposed a novel approach based on Neural Radiance Field, to model scenes across different spectra. 
Thanks to NXDC, we learn an aligned representation across spectra and render images at the same arbitrary resolution from an arbitrary viewpoint, addressing several problems that do arise when attempting to learn a shared NeRF from multiple devices, such as misalignment between sensors and diversity in resolution. Moreover, by learning the relative poses between sensors, we can get rid of cumbersome cross-spectral calibration. We tested X-NeRF on images acquired by our multi spectral rig, showing the effectiveness of our approach in producing high quality registered images with different modalities.
% future works
We believe that our work could be useful in several fascinating applications in multi-modal spectral understanding, which we aim at explore in the future.
Moreover, our study is now limited to forward-facing scene. 
Future research will aim at extending X-NeRF also to 360° scenes, addressing the more challenging lightning conditions occurring and the increased aliasing due to huge difference of resolutions across modalities \cite{Barron_2021_ICCV}.
Finally, another intriguing direction would be to collect images with sensors with extremely different wavelength sensitivity such as X-rays, enabling world understanding at different \textit{3D layers}.

\textbf{Acknowledgements.} We gratefully acknowledge the funding support of Huawei Technologies Oy (Finland).

{\small
\bibliographystyle{ieee_fullname}
\bibliography{egbib}
}

\end{document}